\documentclass[runningheads]{llncs}
\usepackage{graphicx}
\usepackage{tikz}
\usepackage{comment}
\usepackage{amsmath,amssymb}
\usepackage{color}
\usepackage{todonotes}
\usepackage{booktabs}
\usepackage{algorithm}
\usepackage{algorithmic}
\usepackage{subcaption}
\usepackage{bm}

\begin{document}
\pagestyle{headings}
\mainmatter
\def\ECCVSubNumber{7741}

\title{Parameterized Temperature Scaling for Boosting the Expressive Power in Post-Hoc Uncertainty Calibration}

\titlerunning{Parameterized Temperature Scaling}


\author{Christian Tomani\inst{1} \and
Daniel Cremers\inst{1} \and
Florian Buettner\inst{2,3}}

\authorrunning{C. Tomani et al.}

\institute{Technical University of Munich \and
German Cancer Research Center (DKFZ) \and
Goethe University Frankfurt \\
\email{\{christian.tomani, cremers\}@tum.de}\\
\email{florian.buettner@dkfz.de}}

\maketitle

\begin{abstract}
 We address the problem of uncertainty calibration and introduce a novel calibration method,  Parametrized Temperature Scaling (PTS). Standard deep neural networks typically yield uncalibrated predictions, which can be transformed into calibrated confidence scores using post-hoc calibration methods. In this contribution, we demonstrate that the performance of accuracy-preserving state-of-the-art post-hoc calibrators is limited by their intrinsic expressive power. We generalize temperature scaling by computing prediction-specific temperatures, parameterized by a neural network. We show with extensive experiments that our novel accuracy-preserving approach consistently outperforms existing algorithms across a large number of model architectures, datasets and metrics.\footnote{Source code available at: \url{https://github.com/tochris/pts-uncertainty}}
\end{abstract}

\section{Introduction}
Due to their high predictive power, neural network based systems are increasingly used for decision making in real-world applications. Models deployed in such real-world settings, require not only high accuracy, but also reliability and uncertainty-awareness. Especially in safety critical applications such as autonomous driving or in automated factories where average case performance is insufficient, a reliable estimate of the predictive uncertainty of models is crucial. This can be achieved via well-calibrated confidence scores that are representative of the true likelihood of a prediction.

\begin{figure}[!ht]
	\centering
	\begin{subfigure}[t]{0.42\textwidth}
    	\includegraphics[width=\textwidth]{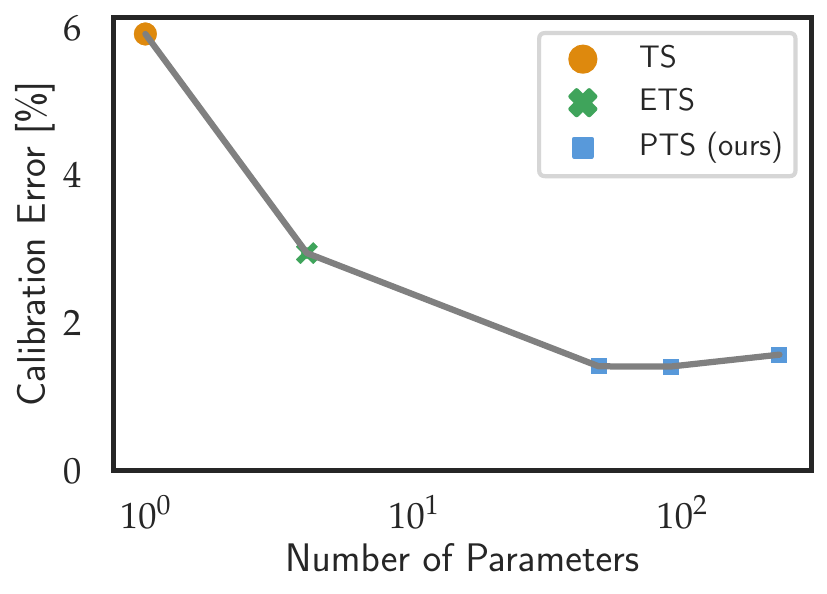}
    	\caption{Increased expressive power of post-hoc methods yields lower calibration error.}
    \end{subfigure}
	\begin{subfigure}[t]{0.48\textwidth}
    	\includegraphics[width=\columnwidth]{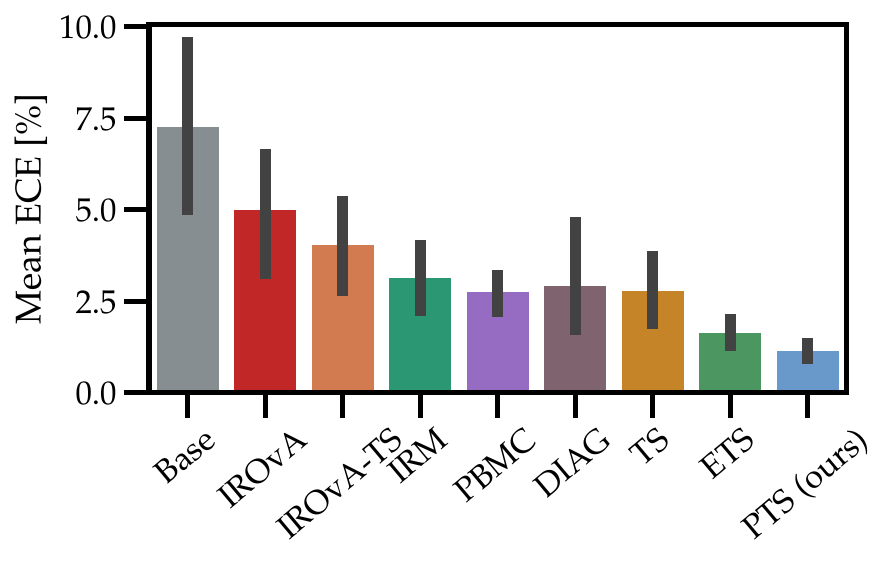}
    	\caption{Our approach improves substantially over baselines across 3 datasets and 9 architectures.}
    \end{subfigure}
\caption{(a) With increased expressive power, temperature scaling-based models yield lower expected calibration errors (ECE). All post-hoc calibration models were optimized to calibrate a MobileNetV2 trained on ImageNet. (b) Bars show the average ECE of all baseline methods. Average is taken over all architectures and baselines and lines indicate standard deviation. PTS improves substantially over all baselines with relative reduction of calibration error of 30\% over ETS and even higher reductions for other baselines.}
\label{fig:ece_expr}
\end{figure}

Since modern neural networks tend to yield systematically overconfident predictions \cite{guo_calibration_2017,minderer2021revisiting}, a number of algorithms for post-hoc calibration have been proposed. These algorithms include parametric approaches that transform the outputs of neural networks based on simple linear models in form of Platt-scaling or Temperature scaling. Alternative non-parametric approaches include  histogram- or regression-based models such as histogram-binning or isotonic regression. Recent research efforts have shown that combining and extending these base techniques \cite{zhang2020mix,kumar2019verified,jang2021improving,ma2021meta} results in a plethora of approaches where no single approach performs best across datasets and model architectures.
Temperature-scaling based approaches are a particularly appealing family of post-hoc calibrators since they do not affect the accuracy of the transformed model and have a high data efficiency, so that they can be applied also in low-data settings with only small validation sets available. However, they are collectively limited by a low expressive power (or model capacity): Temperature scaling \cite{guo_calibration_2017} fits a single scalar parameter, extended temperature scaling \cite{zhang2020mix} is based on a weighted ensemble of 3 fixed temperatures. While non-parametric models are more expressive, they usually do not preserve model accuracy and the accuracy of trained models may decrease substantially after calibration \cite{guo_calibration_2017,zhang2020mix}. Importantly, all temperature-scaling based approaches are based on a fixed calibration map, that transforms all uncalibrated predictions of a neural network into calibrated predictions in the same manner without leveraging information from individual predictions. \\
We hypothesize that the performance of temperature-scaling based post-hoc calibration models is intrinsically limited by their expressive power, which stems from a lack of modeling a prediction-specific transformation. We show that our prediction-specific temperatures are indeed different for each model; in fact, they vary over a wide range of values, which is in stark contrast to only 1 or 3 temperatures for temperature scaling or ensemble temperature scaling and indicates that temperatures calculated based on each prediction separately yield more accurate uncertainty aware post-hoc calibration results.

\subsection{Contributions}
In this work we make the following contributions:

\begin{itemize}
    \item We show that the limiting factor of TS-based post-hoc calibrators is the expressive power of the underlying calibration model.
    \item We generalize temperature scaling based on a highly expressive neural network that computes \textit{prediction-specific} temperatures; we refer to our accuracy-preserving post-hoc calibration approach as Parameterized Temperature Scaling (PTS).
    \item We show that our approach has a similar data-efficiency as state-of-the-art prediction-agnostic post-hoc calibrators.
    \item We demonstrate in exhaustive experiments that PTS outperforms existing methods across a wide range of datasets and models  as "one-size-fits-all" calibrator, without the need to optimize any hyperparameter.
\end{itemize}

\section{Related Work}
In this section, we review existing approaches for post-hoc calibration of trained neural networks.
For this type of post-processing method a validation set, drawn from the generative distribution of the training data, is used to rescale the outputs returned by a trained neural network such that in-domain predictions are well calibrated. Related work can be categorized along two distinct axes, namely parametric vs non-parametric methods and accuracy-preserving methods vs. those where accuracy can change after calibration. While non-parametric approaches tend to have a higher expressive power than parametric models, most non-parametric methods suffer from the drawback that they do not preserve the accuracy of trained neural networks.

\subsection{Non-parametric methods}  A  popular non-parametric post-processing approach is histogram binning \cite{zadrozny2001obtaining}. In brief, all uncalibrated confidence scores $\hat{P}_l$  are partitioned into $M$ bins. Next, a calibrated score $Q_m$ is assigned to each bin by optimizing a bin-wise squared loss.
Extensions to histogram binning include isotonic regression \cite{zadrozny2002transforming} and Bayesian Binning into Quantiles (BBQ) \cite{naeini2015obtaining}. For isotonic regression, uncalibrated confidence scores are divided into $M$ intervals and a piecewise constant function $f$ is fitted on the validation set. This isotonic function is then used to transform uncalibrated outputs into calibrated scores. BBQ is a Bayesian generalization of histogram binning using the concept of Bayesian model averaging. Recently proposed alternatives to histogram-based methods are Gaussian Process based calibration \cite{wenger2020non} and calibration via splines \cite{gupta2020calibration}. While these non-parametric methods do not preserve the accuracy of trained neural networks, Zhang et al. \cite{zhang2020mix} have recently introduced an accuracy-preserving extension of isotonic regression by imposing strict isotonicity on the isotonic function.

\subsection{Parametric methods}
In addition to these non-parametric approaches, also parametric alternatives for post-processing confidence scores exist. For example, the idea of Platt scaling \cite{Platt99probabilisticoutputs} is based on transforming the non-probabilistic outputs (logits) $z_i \in \mathbb{R}$ of a binary classifier to calibrated confidence scores. While initially proposed in the context of support vector machines, Platt scaling has also been used for calibrating other classifiers, including neural networks. More specifically, the logits are transformed to calibrated confidence scores $\hat{Q}_i$ using logistic regression $\hat{Q}_i= \sigma(a z_i + b)$, where $\sigma$ is the sigmoid function. The two parameters $a$ and $b$ are fitted by optimising the negative log-likelihood of the validation set.\\
Guo et al. \cite{guo_calibration_2017} generalized Platt scaling to the multi-class case: Temperature Scaling (TS) is a simple but popular post-processing approach where a scalar parameter $T$ is used to re-scale the logits of a trained neural network. In the case of $C$-class classification, the logits are a $C$-dimensional vector $\mathbf{z}_i \in \mathbb{R}^C$, which are typically transformed into confidence scores $\hat{P}_i$ using the softmax function $\sigma_{SM}$. For temperature scaling, logits are rescaled with temperature $T$ and transformed into calibrated confidence scores $\hat{Q}_i$ using $\sigma_{SM}$ as
\begin{equation}
\hat{Q}_i = \max_c \sigma_{SM}(\mathbf{z}_i/T)^{(c)}
\end{equation}\label{eq:ts}
$T$ is learned by minimizing the negative log-likelihood of the validation set. Zhang et al. \cite{zhang2020mix} have recently introduced an extended temperature scaling, where calibrated predictions are obtained by a weighted sum of predictions re-scaled via three individual temperature terms: an adjustable temperature (as in vanilla temperature scaling), a fixed temperature of 1 and a fixed temperature of $\infty$.
Other generalization of Platt scaling to the multi-class case are vector scaling and matrix scaling \cite{guo_calibration_2017}. Matrix scaling applies a linear transformation $\mathbf{W}\mathbf{z}_i+\mathbf{b}$ to the logits, where $\mathbf{W}$ and $\mathbf{b}$ do not depend on individual predictions. For vector scaling, $\mathbf{W}$ is chosen to be diagonal, so that it can be interpreted as a generalization of Ensemble Temperature Scaling, where each dimension is transformed with its own temperature. In contrast to the non-parametric methods introduced above or other parametric multi-class calibrators such as vector scaling/matrix scaling  or Dirichlet based scaling \cite{milios2018dirichlet},  Temperature Scaling-based methods have the advantage that they do not change the accuracy of the trained neural network. Since re-scaling  does not affect the ranking of the logits, also the maximum of the softmax function remains unchanged. In this work, we build on temperature scaling in order to leverage its accuracy-preserving nature and introduce a generalized formulation that overcomes its limited expressive power.\\
More recently, a family of intra order-preserving function was proposed as post-hoc calibration functions that can preserve the top-k predictions of neural networks \cite{rahimi2020intra}. Confnet \cite{Wan2018ConfnetPW} is another calibration method that obtains better confidence scores by feeding logits into a neural network; however, it is an end-to-end framework optimizing also the weights of the classifier itself.

\begin{figure}[ht]
		\begin{minipage}{\textwidth}
    \begin{algorithm}[H]
    	\caption{Parameterized Temperature Scaling (PTS)\\ \textbf{Input}: Trained classification model $(\hat{Y}, \hat{Z})=h(X)$, validation set $(X, Y)$, initialized calibration network $T = g_\theta(Z)$, number of training steps $S$, batch size $\beta$.}
    	\begin{algorithmic}[1]
    	\FOR{$\varsigma$ in 1:$S$}
        \STATE Read minibatch $MB = (\{X_1 ,\dots, X_\beta\}, \{Y_1 , \dots, Y_\beta\})$ from validation set
        \FOR{$X_b$ in $MB$}
        \STATE Compute calibrated predictions $\sigma_{SM}(\mathbf{z}_b/g_\theta(\mathbf{z}_b^s))^{(c)} $ with $\mathbf{z}_b = h(X_b)$ (eq. \ref{eq:qscore})
        \ENDFOR
        \STATE Compute $L_\theta$ based on $MB$ and do one training step optimizing $\bm{\theta}$ based on $MB$
        \ENDFOR
        \end{algorithmic}
    \end{algorithm}
		\end{minipage}
	\end{figure}


\section{Definitions and problem set-up}
Let $X \in \mathbb{R}^D$ and $Y \in \{1,\dots, C\}$ be random variables that denote the $D$-dimensional input and labels in a classification task with $C$ classes with a ground truth joint distribution $\pi(X, Y ) =
\pi(Y |X)\pi(X)$. The dataset $\mathcal{D}$ consists of $N$ i.i.d. samples $\mathcal{D} = \{(X_n, Y_n)\}_{n=1}^N$ drawn from  $\pi(X, Y )$. Let $h(X) = (\hat{Y},\hat{Z})$ be the output of a trained neural network classifier $h$ predicting a class $\hat{Y}$ and an associated unnormalized logit tupel $\hat{Z}$ based on $X$. $\hat{Z}$ is then transformed into a confidence score $\hat{P}$ associated to $\hat{Y}$ via the softmax function $\sigma_{SM}$ as $\hat{P}=\max_c \sigma_{SM}(\hat{Z})^{(c)}$. In this work, we develop a new approach to improve the quality of the predictive uncertainty of $h$ by improving the calibration of its confidence scores $\hat{P}$.
\paragraph{Uncertainty (miss-)calibration} We define perfect calibration such that accuracy and confidence match for all confidence levels \cite{guo_calibration_2017}:
\begin{eqnarray}
\mathop{\mathbb{P}}(\hat{Y}=Y| \hat{P}=p) = p, \; \; \forall p \in [0,1]
\label{eq:cali}
\end{eqnarray}
Based on equation \ref{eq:cali} it is straight-forward to define miss-calibration as the difference in expectation between confidence and accuracy:
\begin{eqnarray}
\mathop{\mathbb{E}}_{\hat{P}}\left[\big\lvert\mathop{\mathbb{P}}(\hat{Y}=Y| \hat{P}=p) - p\big\rvert \right]
\label{eq:miscal}
\end{eqnarray}

\paragraph{Measuring calibration} The expected calibration error (ECE) \cite{naeini2015obtaining} is a scalar summary measure estimating miss-calibration by approximating equation \ref{eq:miscal} based on predictions, confidence scores and ground truth labels $\{(Y_l,\hat{Y}_l,\hat{P}_l)\}_{l=1}^L$ of a finite number of $L$ samples. ECE is computed by first partitioning all $L$ confidence scores $\hat{P}_l$ into $M$ equally sized bins of size $1/M$ and computing accuracy and average confidence of each bin. Let $B_m$ be the set of indices of samples whose confidence falls into its associated interval $I_m = \left(\frac{m-1}{M} ,\frac{m}{M}\right]$. $\mathrm{conf}(B_m) = 1/|B_m|\sum_{i\in B_m}\hat{P}_i$ and $\mathrm{acc}(B_m) = 1/|B_m|\sum_{i\in B_m} \mathbf{1}(\hat{Y}_i = Y_i)$ are the average confidence and accuracy associated with $B_m$, respectively. The ECE is then computed as
\begin{eqnarray}
\mathrm{ECE}^d = \sum_{m=1}^M \frac{\lvert B_m\rvert}{n}\left\| \mathrm{acc}(B_m) - \mathrm{conf}(B_m)\right\|_d
\end{eqnarray}\label{eq:ece}
with $d$ usually set to 1 for the l1-norm.
While the ECE is the most commonly used measure of miss-calibration, it has some drawbacks. In particular, the choice of bins can result in biased estimates and/or volatility \cite{ashukha2020pitfalls,kumar2019verified,zhang2020mix}. Therefore, alternative formulations to mitigate these issues have been suggested. For example, Zhang et al. \cite{zhang2020mix} have proposed to replace histograms with
non-parametric density estimators and present an ECE based on kernel density estimation (KDE). In addition to top-label ECE (eq. \ref{eq:ece}), class-wise ECE has been proposed as a metric. However, they have been observed to be often contradictory \cite{nixon2019measuring}. Consequently, calibration gain, a dimensionality-independent solution to compare calibration maps was recently introduced \cite{zhang2020mix}. This metric builds on the well-known calibration refinement decomposition \cite{murphy1973new} for the strictly proper scoring loss \cite{gneiting2007strictly}. Orthogonal ways of evaluating calibration include testing a hypothesis of perfect calibration \cite{vaicenavicius2019evaluating}.

\section{Highly expressive post-hoc calibration via parameterized temperature scaling}
To overcome limitations in the expressive power of TS-based methods, we propose to parameterize the temperature in a flexible and expressive manner. Rather then learning a single temperature (or weighted sum of fixed temperatures), we  introduce a dependency of the temperature on the un-normalized logits. In other words, while temperature scaling works by re-scaling any logit tupel of a trained model by the same temperature, PTS introduces a dependency of the temperature on the logit tuple itself. That is, our approach leverages the information present in a logit tupel to compute a prediction-specific temperature.\\
More formally, we propose the following post-hoc calibrator to map unnormalized logits $\mathbf{z}$ to calibrated confidence scores. We start by  parameterizing the temperature $T$ with a flexible neural network as follows:

\begin{eqnarray}\label{eq:PTS}
T(\mathbf{z};\bm{\theta}) = g_\theta(\mathbf{z}^s)
\end{eqnarray}
with $\bm{\theta}$ being the weights of a neural network $g$ parameterizing the scalar temperature $T(\mathbf{z}; \bm{\theta})$ and $\mathbf{z}^s$ being an unnormalised logit tuple sorted by decreasing value.\\
The parameterized temperature is then used to obtain calibrated confidence scores $\hat{Q}_i$ for sample $i$ based on unnormalized logits $\mathbf{z}_i$:
\begin{eqnarray}
\hat{Q}_i(\mathbf{z}_i,\bm{\theta}) &=& \max_c \sigma_{SM}(\mathbf{z}_i/T(\mathbf{z}_i;\bm{\theta}))^{(c)}  \\
& = & \max_c \sigma_{SM}(\mathbf{z}_i/g_\theta(\mathbf{z}_i^s))^{(c)}
\label{eq:qscore}
\end{eqnarray}

We fit a post-hoc calibrator for a trained neural network $h(X)$ by optimizing a squared error loss $L_\theta$ with respect to $\bm{\theta}$.

 \begin{eqnarray}
L_\theta &=& \frac{1}{N} \sum_{n=1}^N
\sum_{c=1}^C(I_{nc}-\sigma_{SM}(\mathbf{z}_i/g_\theta(\mathbf{z}_i^s))^{(c)})^2
\end{eqnarray}\label{eq:lossece}
with $I_{nc}$ being $1$ if sample $n$ has true class $c$, and $0$ otherwise. PTS is summarized in Algorithm 1.\\

Like standard temperature scaling, PTS with a parameterized temperature $T(\mathbf{z};\bm{\theta})$ does not change the accuracy of the trained model since the ranking of the logits remains unchanged.

We first explore the relation between calibration performance and expressive power of a post-hoc calibrator and demonstrate that performance of current state-of-the-art temperature-scaling based calibrators is limited by expressive power.\\
Next, we show that PTS is a one-size-fits-all approach for post-hoc calibration: in contrast to the common state-of the-art where performance varies widely between datasets and network architectures, our approach consistently outperforms state-of-the-art methods on a wide range of datasets and model architectures. We then demonstrate that in spite of the larger number of parameters, PTS has a similar data efficiency compared to low-parametric baselines such as temperature-scaling. Finally, we investigate the dependency structure between logits and their prediction-specific temperature and show that that allowing for non-linearities in this relationship improves calibration performance.

\subsection{Baseline methods and datasets}
With data efficiency and the ability to preserve the trained model's accuracy being key desiderata of post-hoc calibration methods \cite{zhang2020mix}, we mainly focus on accuracy preserving baselines and/or temperature-scaling based methods. We compare our approach to the following baseline methods:

\begin{itemize}
\item Base: Uncalibrated baseline model
\item Temperature scaling (TS): Post-hoc calibration by temperature scaling \cite{guo_calibration_2017}
\item Ensemble Temperature scaling (ETS): Ensemble version of TS with 4 parameters \cite{zhang2020mix}
\item Isotonic regression (IR) \cite{zadrozny2002transforming}
\item Accuracy preserving version of Isotonic regression (IRM) \cite{zhang2020mix}
\item Composite model combining Temperature Scaling and Isotonic Regression (TS-IR) \cite{zhang2020mix}
\item The scaling-binning calibrator, combining temperature scaling with histogram binning (PBMC) \cite{kumar2019verified}
\item Accuracy and intra-order preserving calibration (DIAG) \cite{rahimi2020intra}
\end{itemize}

DIAG can be run with or without hyperparameter optimization; for a fair comparison, since all other baselines including ours have fixed hyperparameters, we report results without hyperparameter optimization.\\
We evaluate the performance of all models on three datasets, namely Imagenet, CIFAR-10 and CIFAR-100. For all datasets, we calibrate various neural network architectures and analyze a mix of complex and less complex settings. To complement the complex architectures needed to perform well on Imagenet and other complex datasets, we explore how our approach (and others) perform in simpler tasks that require less complex models. For Imagenet we used 5 pre-trained models provided as part of tensorflow, namely ResNet50, ResNet152 \cite{he2016deep}, DenseNet169 \cite{huang2017densely}, Xception \cite{chollet2017xception} and MobileNetv2 \cite{sandler2018mobilenetv2}. For CIFAR-10 and CIFAR-100, we trained VGG19 \cite{simonyan2014very} and LeNet5 \cite{lecun1998gradient}.\\

We used a standard setup for evaluating model calibration \cite{guo_calibration_2017} and trained PTS as well as all baselines on the standard validation sets of all datasets. We then evaluated all models by computing the ECE on the standard test sets. For CIFAR-10 and CIFAR-100, we used a validation dataset consisting of  5000 samples and an independent test set of 10000 samples. For ImageNet, we randomly split the hold-out set into a validation set of 12500 samples and a test set of 37500 samples.\\

We quantify the quality of calibration for all experiments using the standard Expected Calibration Error $\mathrm{ECE}^1$ based on 10 bins as well as and Expected Calibration Error based on kernel density estimates, $\mathrm{ECE}_{\mathrm{KDE}}$. In addition, we also report results from the dimensionality-independent calibration gain, which takes all classes into account \cite{zhang2020mix}.\\

PTS was trained as a neural network with 2 fully connected hidden layers with 5 nodes each. Hyperparameters were the same for all experiments, namely a learning rate of 0.00005,  batch size of 1000 and stepsize of 100,000. We further limited $\mathbf{z}^s$ to the top 10 most confident predictions in all settings since we found that they convey sufficient information. That means, PTS can be used in a large variety of settings without the need to optimize hyperparameters.

\begin{table}[t]
\caption{Expected calibration error $\mathrm{ECE^1}$. For all architectures our approach largely outperforms baseline  post-hoc calibrators.}
\centering
\begin{tabular}{lccccccccc}
\toprule
{} &      Base &     IROvA &  IROvA-TS &       IRM &      PBMC &   DIAG&     TS &       ETS &       PTS (ours) \\
\midrule
CIFAR LeNet5         &  1.91 &  1.99 &  1.92 &  1.57 &  2.14 & 1.94&  1.91 &  1.67 &  \textbf{1.47} \\
CIFAR VGG19      &  7.92 &  1.10 &  \textbf{0.77} &  1.03 &  1.66 & 1.51&  1.37 &  1.34 &  0.84 \\
\midrule
CIFAR100 LeNet5      &  7.54 &  1.71 &  2.52 &  3.52 &  2.69 & 3.78&  1.71 &  1.28 &  \textbf{0.70} \\
CIFAR100 VGG19       &  12.96 &  5.73 &  2.88 &  5.28 &  3.05 & 9.30&  3.64 &  2.11 &  \textbf{0.82} \\
\midrule
ImgNet ResNet50    &  6.26 &  6.60 &  5.66 &  2.86 &  3.47 & 1.53&  1.85 &  1.35 &  \textbf{1.34} \\
ImgNet ResNet152   &  6.39 &  6.55 &  5.46 &  2.88 &  3.49 & 1.99&  2.17 &  1.02 &  \textbf{0.97} \\
ImgNet DenNet169 &  6.13 &  6.61 &  5.64 &  2.74 &  3.39 & 1.89&  1.97 &  1.08 &  \textbf{1.05} \\
ImgNet Xception    &  13.25 &  8.31 &  5.43 &  5.34 &  3.26 & -&  4.40 &  1.83 &  \textbf{1.59} \\
ImgNet MobNetV2 &  2.98 &  6.19 &  6.12 &  3.05 &  1.52 & \textbf{1.41}&  5.91 &  2.94 &  1.43 \\
\bottomrule
\end{tabular}
\label{tab:ece_l1}
\end{table}

\section{Experiments and results}

\begin{table*}[!t]
\caption{Expected calibration error $\mathrm{ECE_{KDE}}$. Overall rankings are largely consistent with $\mathrm{ECE^1}$. DIAG did not converge to a meaningful optimum for Xception.}
\centering
\begin{tabular}{lccccccccc}
\toprule
{} &      Base &     IROvA &  IROvA-TS &       IRM &      PBMC &    DIAG & TS &       ETS &       PTS (ours) \\
\midrule
CIFAR LeNet5         &   1.93 &   1.82 &   1.83 &   1.50 &   2.33 & 2.11&  1.97 &   1.82 &   \textbf{1.49} \\
CIFAR VGG19          &   7.34 &   1.24 &   1.11 &   \textbf{1.07} & 1.95&  1.92 &   1.85 &   1.72 &   1.38 \\
\midrule
CIFAR100 LeNet5      &   7.54 &   1.76 &   2.66 &   3.52 &   2.70 & 4.08&  1.73 &   1.07 &   \textbf{0.95} \\
CIFAR100 VGG19       &  12.32 &   5.29 &   2.50 &   5.23 &   3.73 & 9.24&  3.43 &   2.31 &   \textbf{1.05} \\
\midrule
ImgNet ResNet50    &   5.61 &   5.93 &   5.00 &   2.54 &   4.82 & \textbf{1.17}&  1.44 &   1.57 &   1.32 \\
ImgNet ResNet152   &   5.69 &   5.73 &   4.82 &   2.48 &   4.88 & 1.76&  1.85 &   1.39 &   \textbf{0.89} \\
ImgNet DenNet169 &   5.49 &   5.95 &   4.97 &   2.39 &   4.81 & 1.56&  1.53 &   1.29 &   \textbf{1.03} \\
ImgNet Xception    &  12.59 &   7.72 &   4.79 &   4.77 &   3.62 & -&  4.01 &   1.98 &   \textbf{1.26} \\
ImgNet MobNetV2 &   3.10 &   5.75 &   5.63 &   2.93 &   2.09 & 1.55&  5.91 &   3.02 &   \textbf{1.26} \\
\bottomrule
\end{tabular}\label{tab:ece_kde}
\end{table*}

\begin{table*}[!t]
\caption{Calibration Gain (higher is better): Our approach (PTS) largely outperforms baseline  post-hoc calibrators also for a dimensionality-independent calibration metric.}
\centering
\begin{tabular}{lccccccccc}
\toprule
{} &      IROvA &  IROvA-TS &       IRM &      PBMC &    DIAG &    TS &       ETS &       PTS (ours) \\
\midrule
CIFAR LeNet5             	&   0.01  &	0.01  &	\textbf{0.03}  &	-0.05  & 0.00 &	0.01  &	0.02  &	\textbf{0.03}  \\
CIFAR VGG19          	&   0.91  &	\textbf{0.92}  &	0.91  &	0.88  & 0.87 &	0.87  &	0.91  &	0.91  \\
\midrule
CIFAR100 LeNet5          	&   0.69  &	0.65  &	0.48  &	0.59  & 0.55 &	0.67  &	0.70  &	\textbf{0.72}  \\
CIFAR100 VGG19           	&   1.67  &	1.99  &	1.70  &	1.94  & 0.66&	1.92  &	2.03  &	\textbf{2.09}  \\
\midrule
ImgNet ResNet50        	&   -0.02  &	0.14  &	0.39  &	0.26  & \textbf{0.47}&	0.45  &	\textbf{0.47}  &	\textbf{0.47}  \\
ImgNet ResNet152       	&   0.01  &	0.17  &	0.39  &	0.24  & 0.43&	0.42  &	0.46  &	\textbf{0.48}  \\
ImgNet DenNet169     	&   -0.03  &	0.10  &	0.36  &	0.22  & 0.40&	0.39  &	0.42  &	\textbf{0.43}  \\
ImgNet Xception        	&   0.28  &	0.20  &	0.46  &	0.41  & -&	0.33  &	0.44  &	\textbf{0.49}  \\
ImgNet MobNetV2     	&   0.20  &	0.08  &	0.36  &	0.34  & \textbf{0.38}&	0.28  &	0.36  &	\textbf{0.38}  \\
\bottomrule
\end{tabular}\label{tab:ece_calgain}
\end{table*}

\subsection{Higher expressive power leads to better calibration performance}
\begin{figure*}[t!]
	\centering
	\begin{subfigure}[t]{0.45\textwidth}
		\centering
    	\includegraphics[width=\textwidth]{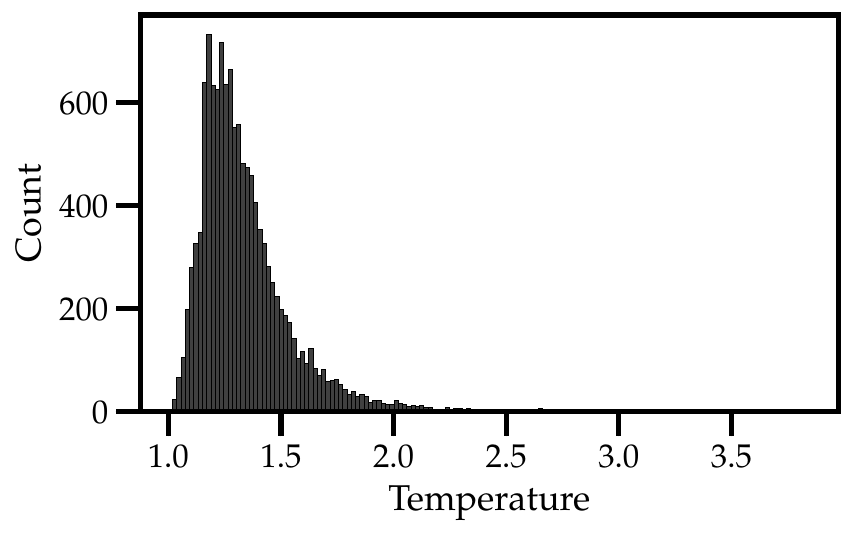}
		\caption{Set of temperatures for DenseNet169.}
	\end{subfigure}
	\centering
	\begin{subfigure}[t]{0.45\textwidth}
		\centering
    	\includegraphics[width=\textwidth]{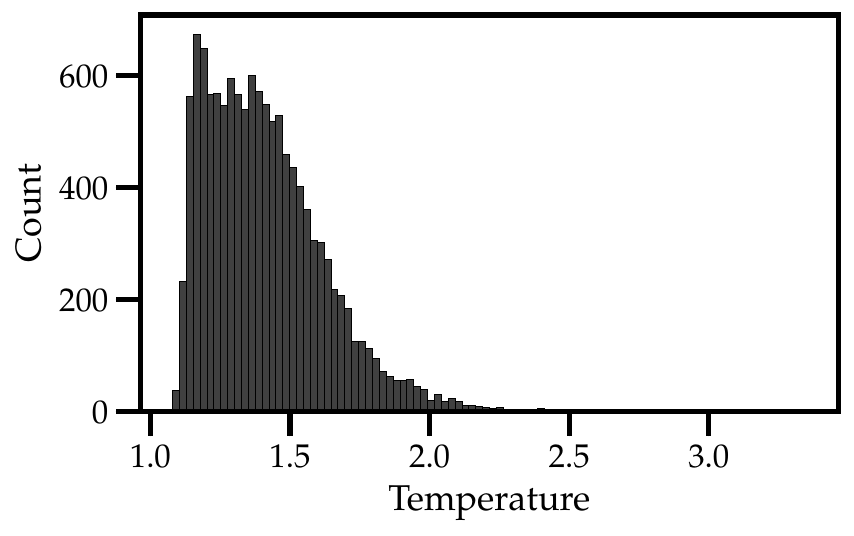}
		\caption{Set of temperatures for ResNet152.}
	\end{subfigure}
	\begin{subfigure}[t]{0.45\textwidth}
		\centering
		\includegraphics[width=\textwidth]{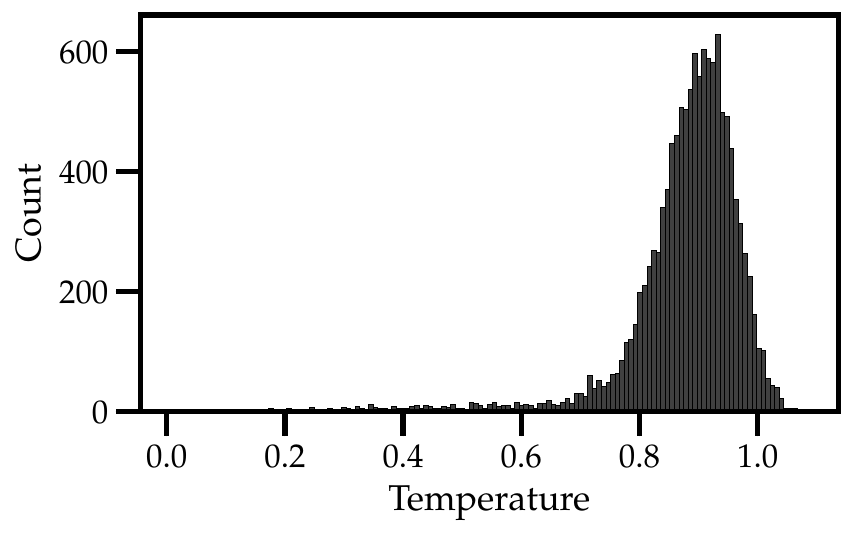}
		\caption{Set of temperatures for MobileNetV2.}
	\end{subfigure}
	\caption{Prediction-specific temperatures inferred by PTS for all samples in the Imagenet test set. For all models, temperatures varied over a wide range of values, which is in stark contrast to only 3 temperatures used in ETS.}
\label{fig:temps}
\end{figure*}

To assess the link between expressive power and calibration for temperature-scaling based models, we train our PTS calibrators with an increasing number of nodes in the hidden layers on the Imagenet validation set. We compare calibration performance of our neural-network-based parameterization of the temperature to post-hoc calibrators with fixed temperature on the Imagenet test set. Fig. 1 illustrates that increasing the expressive power of temperature scaling (based on a single parameter) via a weighted ensembles of 3 temperatures (4 parameters) results in an improved ECE (50\% for MobileNetV2 trained on ImageNet), as previously demonstrated. When further increasing the expressive power of temperature-scaling based calibrators via a neural network, we find that the calibration error further decreases with a larger number of parameters until a plateau is reached (additional improvement in ECE of 52\% for MobileNetV2 trained on ImageNet). We next evaluated the set of temperatures learnt by PTS for predictions from three different models on the Imagenet test set (Fig. \ref{fig:temps}). These temperatures span a wide range of values, which is in stark contrast to only 3 temperatures used in ETS, indicating that for ETS the ensemble of only 3 temperatures limits its calibration performance.\\
Taken together, this illustrates that the performance of conventional temperature-scaling based methods is limited by their inherent expressive power.

\subsection{PTS is a one-size-fits-all post-hoc calibrator}
\begin{figure*}[t!]
	\centering
	\begin{subfigure}[t]{0.45\textwidth}
		\centering
    	\includegraphics[width=\textwidth]{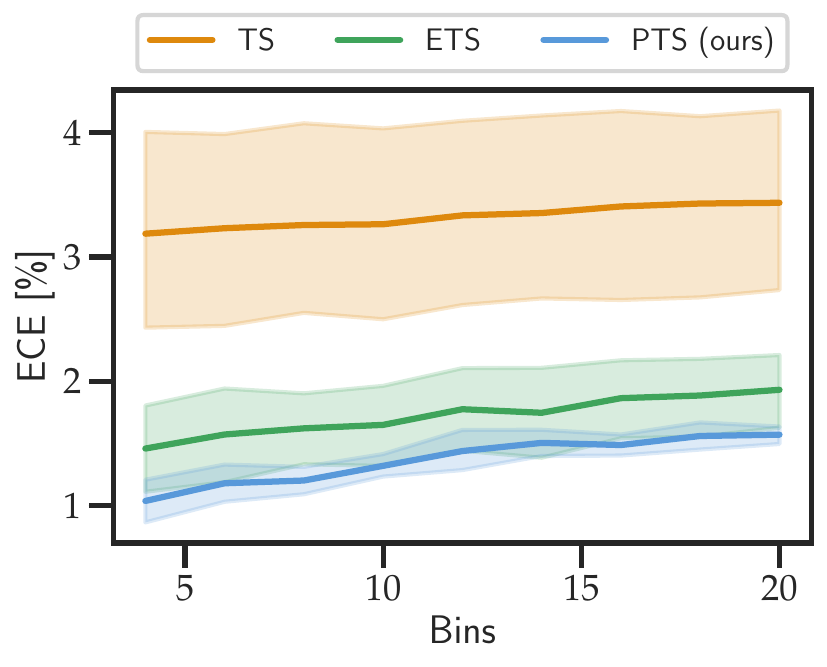}
		\caption{Robustness in terms of number of bins.}
	\end{subfigure}
	\centering
	\begin{subfigure}[t]{0.45\textwidth}
		\centering
    	\includegraphics[width=\textwidth]{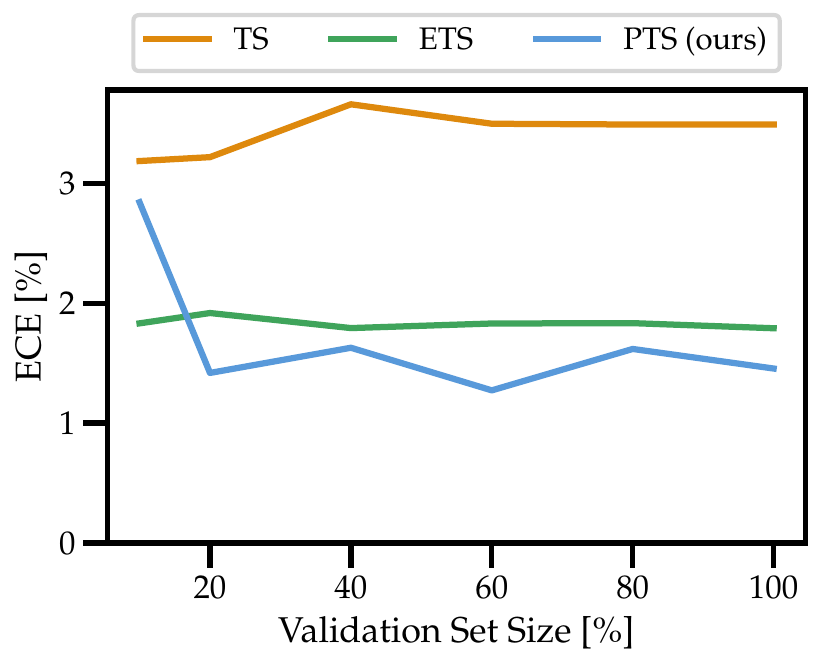}
		\caption{ECE for MobileNetV2 trained on ImageNet for different validation set sizes}
	\end{subfigure}
	\begin{subfigure}[t]{0.45\textwidth}
		\centering
		\includegraphics[width=\textwidth]{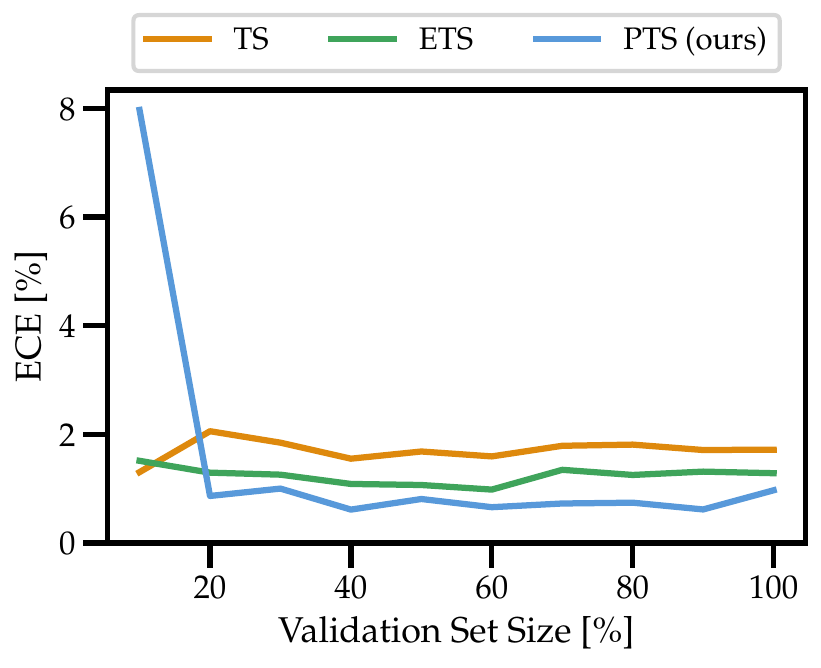}
		\caption{ECE for LeNet5 trained on CIFAR-100 for different validation set sizes}
	\end{subfigure}
	\caption{Robustness of TS-based methods. (a) Mean ECE across 5 architectures trained on ImageNet, with a confidence band illustrating one standard deviation. PTS has the lowest calibration error across all architectures, irrespective of the chosen number of bins. Robustness in terms of dataset size (b and c): ECE for post-hoc calibrators trained on increasingly smaller subsets of the validation sets of ImageNet (b) and CIFAR-100 (c), generated by subsampling decreasing fractions of the full validation set (10\% to 100\%). PTS maintains the high data efficiency inherent in TS methods with low ECE even for small validation sets.}
\label{fig:data_eff}
\end{figure*}

We next evaluate the performance of our approach on a total of 9 deep neural networks trained on CIFAR-10, CIFAR-100 and Imagenet.\\
Table \ref{tab:ece_l1} shows the standard ECE, Table \ref{tab:ece_kde} the KDE-based ECE and Table \ref{tab:ece_calgain} calibration gain for all experiments. Rankings for all metrics are largely consistent and our approach outperforms baselines in all settings in terms of $\mathrm{ECE}^1$. $\mathrm{ECE_{KDE}}$ suggests that in settings with low complexity  - i.e. a simple dataset with low number of classes such as CIFAR-10 and/or a simple architecture such as LeNet - performance of PTS is comparable to ETS only. The more complex a setting, the larger the gain of PTS. This range of complex architectures and datasets is particularly relevant in practice since Guo et al. \cite{guo_calibration_2017} have shown that it is particularly modern architectures that are prone to mis-calibration. Additionally, Table \ref{tab:ece_calgain} shows the calibration gain \cite{zhang2020mix}, which indicates that PTS yields an improved performance even for metrics, which are dimensionality-independent.\\
To make sure the choice of bin size when computing the ECE does not affect our findings, we computed $\mathrm{ECE}^1$ for bin sizes $M$ ranging from 5 to 20 in steps of 2. Figure \ref{fig:data_eff} (a) illustrates the mean ECE across the 5 architectures trained on ImageNet and calibrated using TS, ETS and PTS. While small bin sizes result in a systematically smaller ECE - a known bias \cite{kumar2019verified} - PTS outperforms the other TS-based methods for all bin sizes with rankings being unchanged.\\


\subsection{PTS is data-efficient}
A major advantage of TS-based models over other approaches is their high data-efficiency paired with accuracy-preserving properties. We therefore designed experiments to quantify the data-efficiency of PTS. To this end, we fitted our model on increasingly smaller subsets of the validation set to calibrate a MobileNetV2 architecture trained on ImageNet and a LeNet5 architecture trained on CIFAR-100. In both cases we varied the size of the subsets from 10\% to 100\% of the respective standard validation set size.  When evaluating ECE on the test set, we found that like vanilla TS and ETS, our model yielded excellent performance even when trained on small fractions of the validation set, maintaining one of the key advantages of TS-based models (Figure \ref{fig:data_eff} (b) and (c)). We confirmed the robustness of these findings, by performing the dataset-size vs ECE experiment for all other models trained on imagenet. We summarized the results by normalizing ECE relative to PTS for each dataset size and model. We then took the average across all models. Table \ref{tab:val2} shows that across all models, only for very small dataset sizes of 10\%, PTS is outperformed by other approaches.

\begin{figure*}[t!]
	\centering
	\begin{subfigure}[t]{0.45\textwidth}
		\centering
    	\includegraphics[width=\textwidth]{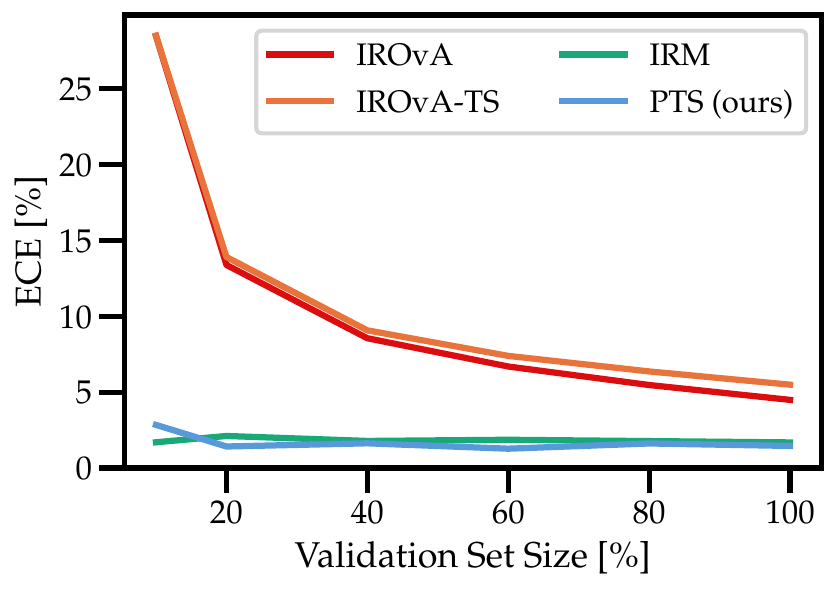}
		\caption{ECE for MobileNetV2 trained on ImageNet}
	\end{subfigure}
	\begin{subfigure}[t]{0.45\textwidth}
		\centering
		\includegraphics[width=\textwidth]{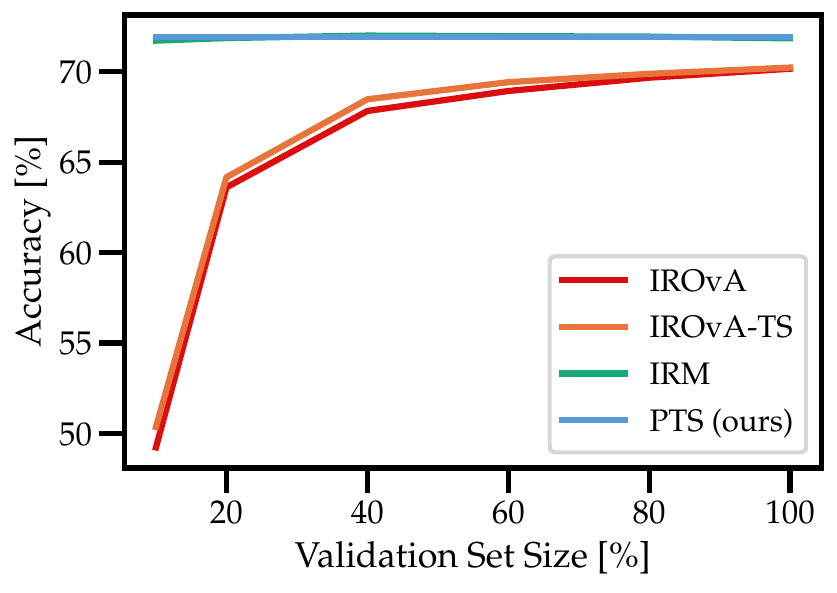}
		\caption{Accuracy for MobileNetV2 trained on ImageNet}
	\end{subfigure}

	\caption{ECE and accuracy for non-parametric post-hoc calibrators on validation sets of decreasing size. Non-parametric calibration methods suffer from a low data-efficiency and decrease in accuracy compared to PTS.}
\label{fig:data_eff_nonparam}
\end{figure*}

\begin{table}
    \centering
    \caption{Normalized ECE relative to PTS for 5 Imagenet models across different validation set sizes.}
\begin{tabular}{lccccc}
\toprule
Validation Set Size [$\%$] &      IROvA &      IROvA-TS  &  IRM  & TS & ETS\\
\midrule
10 & 6.84 & 6.81 & 0.83 & 0.71 & 0.40\\
20 & 11.92 & 11.42    & 2.62 & 2.12 & 1.11\\
40 & 7.85  & 7.14 & 2.50 & 2.14 & 1.13\\
60 & 6.17  & 5.59 & 2.16 & 1.92 & 1.00\\
80 & 5.96  & 5.09 & 2.47 & 2.13 & 1.00\\
100 & 5.67  & 4.81 & 2.63 & 2.28 & 1.16\\
\bottomrule
\end{tabular}
\label{tab:val2}
\end{table}
These findings are in contrast to non-parametric models, which also tend to have a higher expressive power than standard TS approaches. When repeating the data-efficiency experiment for this family of models, we found that calibration error increased substantially with decreasing validation set size. In addition, IROVA and IROVA-TS also yielded a substantial decrease in accuracy (Figure \ref{fig:data_eff_nonparam}).

\subsection{The relationship between logits and temperature is non-linear}
 We finally assessed whether the information present in the logits can be captured via a linear model or whether non-linearities need to be accounted for. To this end, we trained a linear version of our neural network without non-linear activation function (with the same number of parameters as PTS to ensure a fair comparison) as calibration map. Using this linear model to calibrate trained networks with all assessed architectures on Imagenet results on average in a substantially higher ECE of 1.36 compared to 1.28 using a nonlinear model (Table \ref{tab:lin}).
This indicated that it is not sufficient to generate prediction-specific temperatures using a linear model, but that the relationship between logits and temperature is non-linear.

\begin{table}[!h]
    \centering
    \caption{ECE for linear and non-linear models on Imagenet. Indicates that generating temperatures using simply a linear model is not sufficient.}
\begin{tabular}{lcc}
\toprule
{} &      Linear Model &      PTS (ours) \\
\midrule
ImgNet ResNet50    & 1.47  & \textbf{1.34}       \\
ImgNet ResNet152   & 1.02  & \textbf{0.97}       \\
ImgNet DenNet169 & 1.22  & \textbf{1.05}       \\
ImgNet Xception    & 1.62  & \textbf{1.59}       \\
ImgNet MobNetV2 & 1.49  & \textbf{1.43}       \\
\bottomrule
\end{tabular}
\label{tab:lin}
\end{table}


\section{Conclusion}
In this work, we have introduced a novel approach for accuracy-preserving post-hoc calibration by modeling prediction-specific temperatures. To boost the expressive power of TS-based models we introduce a dependency of the temperature on the predicted logits and propose a parameterization of the temperature itself using a neural network. These prediction-specific temperatures make up a highly expressive, accuracy-preserving and data-efficient generalization of temperature scaling. In extensive experiments, we show that this approach results in substantially lower calibration errors than existing post-hoc calibration approaches, with an average improvement over the current state-of-the-art (ETS) of 30\%.\\

\paragraph{Acknowledgements:}
This work was supported by the Munich Center for Machine Learning and has been funded by the German Federal Ministry of Education
and Research (BMBF) under Grant No. 01IS18036B.

\bibliographystyle{splncs04}
\bibliography{references}

\end{document}